\documentclass[11pt]{article}
\usepackage{acl2014}
\usepackage{times}
\usepackage{url}
\usepackage{latexsym}

\usepackage{amsmath}
\usepackage{amsfonts}

\usepackage{textcomp}
\usepackage{color}
\usepackage{fancyvrb}
\usepackage{graphicx}
\usepackage{fancyvrb}
\usepackage{multirow}
\usepackage{tabularx}
\usepackage{umoline}
\usepackage{caption}
\usepackage{subcaption}

\definecolor{green}{rgb}{0.1,0.5,0.1}
\definecolor{red}{rgb}{1.0,0.2,0.2}
\definecolor{blue}{rgb}{0.0,0.0,1.0}

\newcommand{\com}[1]{}
\newcommand{\add}[1]{{#1}}

\title{Molding CNNs for text: non-linear, non-consecutive convolutions} 

\author{Tao Lei, Regina Barzilay, and Tommi Jaakkola\\
    Computer Science and Artificial Intelligence Laboratory\\
    Massachusetts Institute of Technology\\
    {\tt \{taolei, regina, tommi\}@csail.mit.edu} \\}

\date{}

\begin{document}
\maketitle

\begin{abstract}
The success of deep learning often derives from well-chosen operational building blocks. In this work, we revise the temporal convolution operation in CNNs to better adapt it to text processing. Instead of concatenating word representations, we appeal to tensor algebra and use low-rank n-gram tensors to directly exploit interactions between words already at the convolution stage. Moreover, we extend the n-gram convolution to non-consecutive words to recognize patterns with intervening
words. Through a combination of low-rank tensors, and pattern weighting, we can efficiently evaluate the resulting convolution operation via dynamic programming. We test the resulting architecture on standard sentiment classification and news categorization tasks. Our model achieves state-of-the-art performance both in terms of accuracy and training speed. For instance, we obtain 51.2\% accuracy on the fine-grained sentiment classification task.\footnote{Our code and data are available at
\url{https://github.com/taolei87/text_convnet}}
\end{abstract}

\section{Introduction}

Deep learning methods and convolutional neural networks (CNNs) among them have
become de facto top performing techniques across a range of NLP tasks such as
sentiment classification, question-answering, and semantic parsing. As methods,
they require only limited domain knowledge to reach respectable performance with
increasing data and computation, yet permit easy architectural and operational
variations so as to fine tune them to specific applications to reach top
performance. Indeed, their success is often contingent on specific architectural
and operational choices. 

CNNs for text applications make use of temporal convolution operators or
filters. Similar to image processing, they are applied at multiple
resolutions, interspersed with non-linearities and pooling. The convolution
operation itself is a \emph{linear} mapping over ``n-gram vectors'' obtained by
concatenating \emph{consecutive} word (or character) representations. We argue
that this basic building block can be improved in two important respects. First,
the power of n-grams derives precisely from multi-way interactions and these are
clearly missed (initially) with linear operations on stacked n-gram vectors.
Non-linear interactions within a local context have been shown to improve
empirical performance in various
tasks~\cite{mitchell2008vector,kartsaklis2012unified,socher2013sentiment}.
Second, many useful patterns are expressed as \emph{non-consecutive} phrases,
such as semantically close multi-word expressions (e.g.,``\underline{not} that
\underline{good}'', ``\underline{not} nearly as \underline{good}''). In typical
CNNs, such expressions would have to come together and emerge as useful patterns
after several layers of processing. 

We propose to use a feature mapping operation based on \emph{tensor products} instead
of linear operations on stacked vectors. This enables us to directly tap into
non-linear interactions between adjacent word feature
vectors~\cite{socher2013sentiment,Lei14}. To offset the accompanying parametric explosion we maintain a low-rank representation of the tensor parameters. Moreover, we show that this feature mapping can be applied to all possible \emph{non-consecutive} n-grams in the sequence with an exponentially decaying weight depending on the length of the span. Owing to the low rank representation of the tensor, this operation can be performed efficiently in linear time with respect to the sequence length via dynamic programming. Similar to traditional convolution operations, our non-linear feature mapping can be applied successively at multiple levels.

We evaluate the proposed architecture in the context of sentence sentiment
classification and news categorization. On the Stanford Sentiment Treebank
dataset, our model obtains state-of-the-art performance among a variety of
neural networks in terms of both accuracy and training cost. \add{Our model achieves
51.2\% accuracy on fine-grained classification and 88.6\% on binary
classification, outperforming the best published numbers obtained by a deep
recursive model~\cite{tai2015improved} and a convolutional model~\cite{Kim14}.
On the Chinese news categorization task, our model achieves 80.0\% accuracy,
while the closest baseline achieves 79.2\%.}

\section{Related Work}

Deep neural networks have recently brought about significant advancements in
various natural language processing tasks, such as
language modeling~\cite{bengio2003neural,mikolov2010recurrent}, sentiment
analysis~\cite{socher2013sentiment,iyyer2015,le2015compositional}, syntactic
parsing~\cite{Collobert:2008,SocherEtAl2011:RNN,chen2014fast} and machine
translation~\cite{bahdanau2014neural,devlin2014fast,sutskever2014sequence}.  
Models applied in these tasks exhibit significant architectural differences,
ranging from recurrent neural networks~\cite{mikolov2010recurrent,kalchbrennerB13} to
recursive models~\cite{pollack1990recursive,kuchler1996inductive}, and including
convolutional neural nets~\cite{Collobert:2008,collobert2011natural,yih2014semantic,shen2014learning,kalchbrenner2014,zhang2015text}. 

Our model most closely relates to the latter. Since these models have originally been developed for computer vision~\cite{lecun-98}, their application to NLP tasks introduced a number of modifications. For instance, \newcite{collobert2011natural} use the max-over-time pooling operation to aggregate the features over the input sequence. This variant has been successfully applied to semantic parsing~\cite{yih2014semantic} and information
retrieval~\cite{shen2014learning,gao2014modeling}. \newcite{kalchbrenner2014} instead propose (dynamic) k-max pooling operation for modeling sentences. In addition, \newcite{Kim14} combines CNNs of different filter widths and either static or fine-tuned word vectors. In contrast to the traditional CNN models, our method considers non-consecutive n-grams thereby expanding the representation capacity of the model. Moreover, our model captures non-linear interactions
within n-gram snippets through the use of tensors, moving beyond direct linear projection operator used in standard CNNs.  As our experiments demonstrate these advancements result in improved performance.

\section{Background}

\newcommand{\seq}{\mathbf{x}}

Let $\seq \in \mathbb{R}^{L\times d}$ be the input sequence such as a document or sentence. Here $L$ is the length of the sequence and each $\seq_i\in \mathbb{R}^{d}$ is a vector representing the $i^{th}$ word. The (consecutive) \emph{n-gram}
vector ending at position $j$ is obtained by simply concatenating the corresponding word vectors
\begin{align*}
v_j \ =\ [\seq_{j-n+1};\seq_{j-n+2};\cdots; \seq_{j}]
\end{align*}
Out-of-index words are simply set to all zeros.

The traditional convolution operator is parameterized by filter matrix
$\mathbf{m} \in \mathbb{R}^{nd\times h}$ which can be thought of as 
$n$ smaller filter matrices applied to each $\seq_{i}$ in
vector $v_j$. The operator maps each n-gram vector $v_j$ in the input sequence to $\mathbf{m}^{\top} v_j\ \in\
\mathbb{R}^h$ so that the input
sequence $\seq$ is transformed into a sequence of feature representations,
\begin{align*}
\left[ \mathbf{m}^\top
v_1, \cdots, \mathbf{m}^\top v_L \right] \quad\in\quad \mathbb{R}^{L\times h}
\end{align*}
The resulting feature values are often passed through non-linearities such as the hyper-tangent (element-wise) as well as aggregated or reduced by ``sum-over'' or ``max-pooling'' operations for later (similar stages) of processing.

The overall architecture can be easily modified by replacing the basic n-gram vectors and the convolution operation with other feature mappings. Indeed, we appeal to tensor algebra to introduce a non-linear feature mapping that operates on non-consecutive n-grams.

\section{Model}

\paragraph{N-gram tensor} Typical $n-$gram feature mappings where concatenated word vectors are mapped linearly to feature coordinates may be insufficient to directly capture relevant
information in the $n-$gram. As a remedy, we replace concatenation with a tensor product. Consider a 3-gram $(\seq_1, \seq_2, \seq_3)$ and the corresponding tensor product $\seq_1\otimes \seq_2\otimes \seq_3$. The tensor product is a 3-way array of coordinate interactions such that each $ijk$ entry of the tensor is given by the product of the corresponding coordinates of the word vectors 
\begin{align*}
\left(\seq_1\otimes \seq_2\otimes \seq_3\right)_{ijk} =
\seq_{1i}\cdot\seq_{2j}\cdot \seq_{3k}
\end{align*}
Here $\otimes$ denotes the tensor product operator. The tensor product of a 2-gram analogously gives a two-way array or matrix $\seq_1\otimes \seq_2\ \in\ \mathbb{R}^{d\times d}$. The n-gram tensor can be seen as a direct generalization of the typical concatenated vector\footnote{To see this, consider word vectors with a ``bias'' term ${\seq_i}' = [ \seq_i ; 1 ]$. The tensor product of n such vectors includes the concatenated vector as a subset of tensor entries but, in addition, contains all up to $n^\text{th}$-order interaction terms.}.

\paragraph{Tensor-based feature mapping} Since each n-gram in the sequence is now expanded into a high-dimensional tensor using tensor products, the set of filters are analogously maintained as high-order tensors. In other words, our filters are  linear mappings over the higher dimensional interaction terms rather than the original word coordinates. 

\newcommand{\filters}{\mathbf{T}}

Consider again mapping a 3-gram $(\seq_1, \seq_2, \seq_3)$ into a feature
representation. Each filter is a 3-way tensor with dimensions $d\times d\times d$. The set of $h$ filters, denoted as $T$, is a 4-way tensor of
dimension $d\times d\times d\times h$, where each $d^3$ slice of $T$ represents a single filter and $h$ is the number of such filters, i.e., the feature dimension.
The resulting $h-$dimensional feature representation $z \in \mathbb{R}^h$ for the 3-gram $(\seq_1, \seq_2, \seq_3)$ is obtained by multiplying the filter $\filters$ and the 3-gram tensor as follows. The $l^{th}$ coordinate of $z$ is given by
\begin{align}
z_l &= \sum_{ijk} \filters_{ijkl} \cdot \left(\seq_1\otimes \seq_2\otimes
\seq_3\right)_{ijk} \nonumber \\
    &= \sum_{ijk} \filters_{ijkl} \cdot \seq_{1i} \cdot \seq_{2j} \cdot
    \seq_{3k} \label{eq:map1}
\end{align}
The formula is equivalent to summing over all the third-order polynomial interaction terms where tensor $\filters$ stores the coefficients. 

Directly maintaining the filters as full tensors leads to parametric explosion.   Indeed, the size of the tensor $\filters$ (i.e.
$h\times d^n$) would be too large even for typical low-dimensional word vectors where, e.g., $d=300$. To this end, we assume a
\emph{low-rank factorization} of the tensor $\filters$, represented in the Kruskal form. Specifically, $\filters$ is decomposed into a sum of $h$ rank-1 tensors
\begin{align*}
\filters = \sum_{i=1}^h \mathbf{P}_i\otimes \mathbf{Q}_i \otimes \mathbf{R}_i
\otimes \mathbf{O}_i
\end{align*}
where $\mathbf{P}, \mathbf{Q}, \mathbf{R} \in \mathbb{R}^{h\times d}$ and
$\mathbf{O} \in \mathbb{R}^{h\times h}$ are four smaller parameter matrices.
$\mathbf{P}_i$ (similarly $\mathbf{Q}_i$, $\mathbf{R}_i$ and $\mathbf{O}_i$)
denotes the $i^{th}$ row of the matrix. Note that, for simplicity, we have assumed that the number of rank-1 components in the decomposition is equal to the feature dimension $h$. Plugging the
low-rank factorization into Eq.(\ref{eq:map1}), the feature-mapping can be rewritten in a vector form as
\begin{align}
z\ =\ \mathbf{O}^\top \left( \mathbf{P}\seq_1 \odot \mathbf{Q}\seq_2 \odot
\mathbf{R}\seq_3 \right) \label{eq:map2}
\end{align}
where $\odot$ is the element-wise product such that, e.g., $(a\odot b)_k = a_k\times
b_k$ for $a,b\in \mathbb{R}^m$. Note that while $\mathbf{P}\seq_1$ (similarly $\mathbf{Q}\seq_2$ and $\mathbf{R}\seq_3$) is a linear mapping from each word $\seq_1$ (similarly $\seq_2$ and $\seq_3$) into a $h$-dimensional feature space, higher order terms arise from the element-wise products.

\paragraph{Non-consecutive n-gram features} Traditional convolution uses consecutive n-grams in the feature map.
Non-consecutive n-grams may nevertheless be helpful since phrases such as ``\underline{not} \underline{good}'',
``\underline{not} so \underline{good}'' and ``\underline{not} nearly as
\underline{good}'' express similar sentiments but involve variable
spacings between the key words. Variable spacings are not effectively captured by fixed n-grams. 

We apply the feature-mapping in a weighted manner to all n-grams thereby gaining access to patterns such as
``\underline{not} ...
\underline{good}''. Let $z[i,j,k]\in \mathbb{R}^h$ denote the feature representation corresponding to a 3-gram $(\seq_i, \seq_j, \seq_k)$ of words in positions $i$, $j$, and $k$ along the sequence. This vector is calculated analogously to Eq.(\ref{eq:map2}),
\begin{align*}
z[i,j,k]\ =\ \mathbf{O}^\top
\left(\mathbf{P}\seq_i\odot\mathbf{Q}\seq_j\odot\mathbf{R}\seq_k\right)
\end{align*}
We will aggregate these vectors into an $h-$dimensional feature representation at each position in the sequence. The idea is similar to neural
bag-of-words models where the feature representation for a document or
sentence is obtained by averaging (or summing) of all the word vectors. In our case, we define the aggregate representation $z_3[k]$ in position $k$ as the weighted sum of all 3-gram feature representations ending at position $k$, i.e.,
\begin{align}
z_3[k] &= \sum_{i<j<k} z[i,j,k] \cdot \lambda^{(k-j-1)+(j-i-1)} \nonumber \\
       &= \sum_{i<j<k} z[i,j,k] \cdot \lambda^{k-i-2} \label{eq:f3}
\end{align}
where $\lambda\in [0,1)$ is a decay factor that down-weights 3-grams with longer
spans (i.e., 3-grams that skip more in-between words). As $\lambda\rightarrow 0$ all non-consecutive 3-grams are omitted, $z_3[k]=z[k-2,k-1,k]$, and the model acts like a traditional model with only consecutive n-grams. When $\lambda > 0$, however, $z_3[k]$ is a weighted average of many 3-grams with variable spans.

\begin{figure*}[th]
\centering
\includegraphics[height=2.3in]{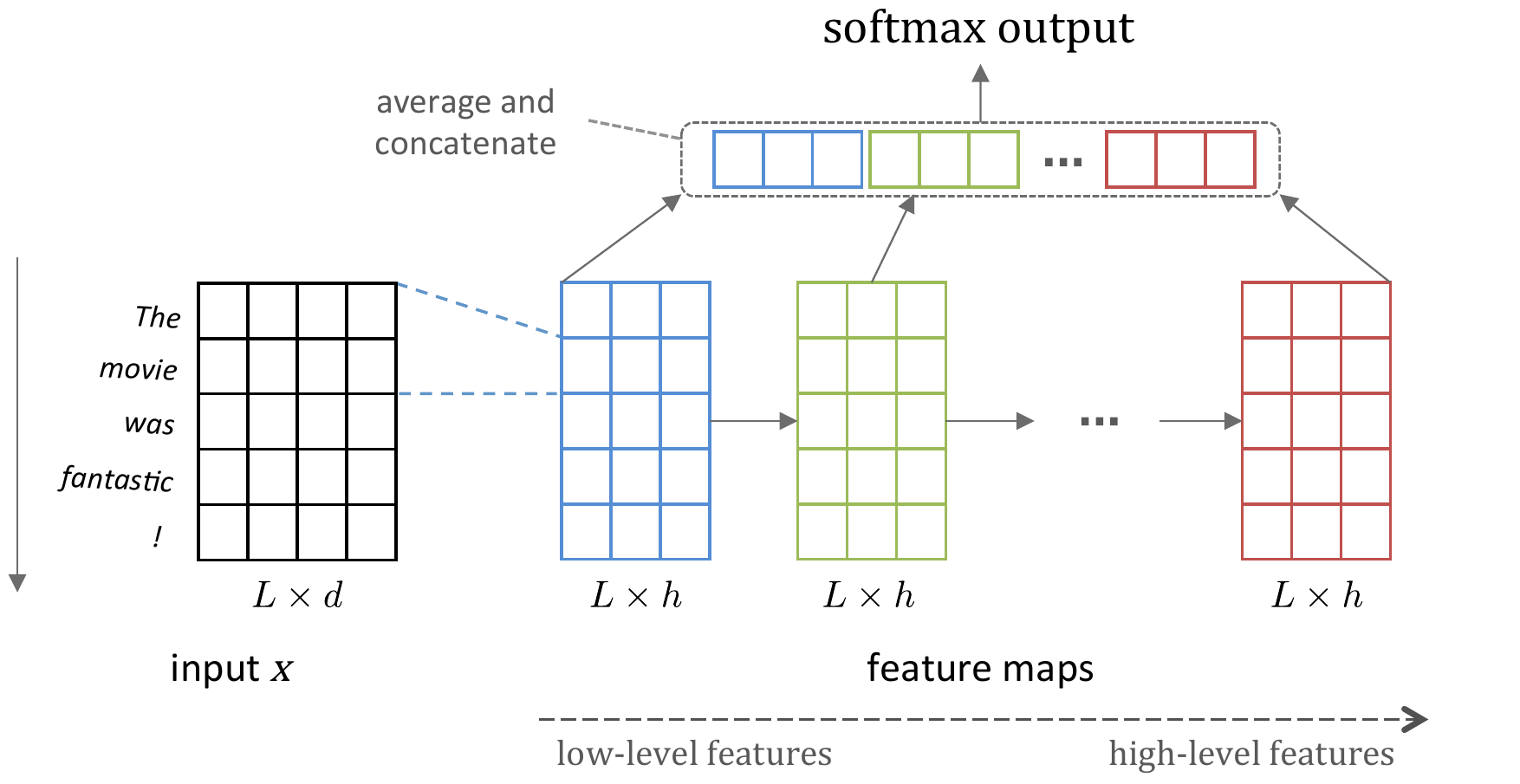}
\vspace{-0.05in}
\caption{Illustration of the model architecture. The input is represented as a
matrix where each row is a d-dimensional word vector. Several feature map layers
(as described in Section 4) are stacked, mapping the input into different levels of
feature representations. The features are averaged within each layer and then
concatenated. Finally a softmax layer is applied to obtain the prediction output.}
\label{figure:model}
\end{figure*}

\paragraph{Aggregating features via dynamic programming} Directly calculating
$z_3[\cdot]$ according to Eq.(\ref{eq:f3}) by enumerating all 3-grams would require $O(L^3)$ feature-mapping operations. We can, however, evaluate the features more efficiently by relying on the associative and distributive properties of the feature operation in Eq.(\ref{eq:map2}). 

Let $f_3[k]$ be a dynamic programming table representing the sum
of 3-gram feature representations before multiplying with matrix $\mathbf{O}$.
That is, $z_3[k] = \mathbf{O}^\top f_3[k]$ or, equivalently,
\begin{align*}
f_3[k] &= \sum_{i<j<k} \lambda^{k-i-2} \cdot \left(\mathbf{P}\seq_i\odot\mathbf{Q}\seq_j\odot\mathbf{R}\seq_k\right)
\end{align*}
We can analogously define $f_1[i]$ and $f_2[j]$ for
1-grams and 2-grams,
\begin{align*}
f_1[i] &= \mathbf{P}\seq_i \\
f_2[j] &= \sum_{i<j}\ \lambda^{j-i-1} \cdot
\left(\mathbf{P}\seq_i\odot\mathbf{Q}\seq_j\right)
\end{align*}
These dynamic programming tables can be calculated recursively according to the following formulas:
\begin{align*}
f_1[i] &=\ \mathbf{P}\seq_i \\
s_1[i] &=\ \lambda\cdot s_1[i-1] + f_1[i]\quad\quad\quad \\
\\
f_2[j] &=\ s_1[j-1] \odot \mathbf{Q}\seq_j \\
s_2[j] &=\ \lambda\cdot s_2[j-1] + f_2[j] \\
\\
f_3[k] &=\ s_2[k-1] \odot \mathbf{R}\seq_k\\
\\
z[k]  &= \mathbf{O}^\top \left(f_1[k]+f_2[k]+f_3[k]\right)
\end{align*}
where $s_1[\cdot]$ and $s_2[\cdot]$ are two auxiliary tables. The resulting $z[\cdot]$ is the sum of 1, 2, and 3-gram features. We found that aggregating
the 1,2 and 3-gram features in this manner works better than using 3-gram features alone. Overall, the n-gram feature aggregation can be performed in $O(Ln)$ matrix
multiplication/addition operations, and remains linear in the sequence
length.

\paragraph{The overall architecture} The dynamic
programming algorithm described above maps the original input sequence to a sequence of feature representations $\mathbf{z}=z[1:L] \in \mathbb{R}^{L\times h}$. As in standard convolutional architectures, the resulting sequence can be used in multiple ways. One can directly aggregate it to a classifier or expose it to non-linear element-wise transformations and use it as an input to another sequence-to-sequence feature mapping. 

The simplest strategy (adopted in neural bag-of-words models) would be to average the feature representations and pass the resulting averaged vector directly to a softmax output unit
\begin{align*}
\bar{z} &= \frac{1}{L} \sum_{i=1}^L z[i] \\
\tilde{y} &= \text{softmax}\left(\mathbf{W}^\top \bar{z}\right)
\end{align*}
Our architecture, as illustrated in Figure~\ref{figure:model}, includes two additional refinements. First, we add a non-linear activation function
after each feature representation, i.e. $\mathbf{z'} =
\text{ReLU}\left(\mathbf{z}+b\right)$, where $b$ is a bias vector and
$\text{ReLU}$ is the rectified linear unit function. Second, we stack multiple tensor-based feature mapping layers. That is, the input
sequence $\seq$ is first processed into a feature sequence and passed through the non-linear transformation to obtain $\mathbf{z}^{(1)}$. The resulting feature sequence $\mathbf{z}^{(1)}$ is then analogously processed by another layer, parameterized by a different set of feature-mapping matrices $\mathbf{P},\cdots,\mathbf{O}$, to obtain a higher-level
feature sequence $\mathbf{z}^{(2)}$, and so on. The output feature
representations of all these layers are averaged within each layer and concatenated as shown in Figure~\ref{figure:model}. The final
prediction is therefore obtained on the basis of features across the levels.

\section{Learning the Model}

\newcommand{\loss}{\text{loss}}

Following standard practices, we train our model by minimizing the cross-entropy
error on a given training set. For a single training sequence $\seq$ and the
corresponding gold label $y\in [0,1]^m$, the error is defined as,
\begin{align*}
\loss\left(\seq, y\right) = \sum_{l=1}^m y_l \log\left(\tilde{y}_l\right)
\end{align*}
where $m$ is the number of possible output label.

The set of model parameters (e.g. $\mathbf{P}, \cdots, \mathbf{O}$ in each
layer) are updated via stochastic gradient descent using AdaGrad
algorithm~\cite{adagrad}.

\paragraph{Initialization} We initialize matrices $\mathbf{P}, \mathbf{Q},
\mathbf{R}$ from uniform distribution $\left[ -\sqrt{3/d}, \sqrt{3/d} \right]$
and similarly $\mathbf{O} \sim \text{U}\left[ -\sqrt{3/h}, \sqrt{3/h} \right]$.
In this way, each row of the matrices is an unit vector in expectation, and each
rank-1 filter slice has unit variance as well,
\begin{align*}
\mathbb{E}\left[
\|\mathbf{P}_i\otimes\mathbf{Q}_i\otimes\mathbf{R}_i\otimes\mathbf{O}_i\|^2
\right] = 1
\end{align*}
In addition, the parameter matrix $\mathbf{W}$ in the softmax output layer is
initialized as zeros, and the bias vectors $b$ for $\text{ReLU}$ activation
units are initialized to a small positive constant $0.01$.

\paragraph{Regularization} We apply two common techniques to avoid overfitting
during training. First, we add L2 regularization to all parameter values with
the same regularization weight. In addition, we randomly dropout~\cite{dropout}
units on the output feature representations $\mathbf{z}^{(i)}$ at each level.

\section{Experimental Setup}

\begin{table*}[th!]
\centering
\begin{tabular}{l|cc|cc|c|c}
\hline
\multirow{2}{*}{\textbf{Model}} & \multicolumn{2}{c|}{\textbf{Fine-grained}} & \multicolumn{2}{c|}{\textbf{Binary}} & \multicolumn{2}{c}{\textbf{Time} (in seconds)} \\
\cline{2-7}
 & {~~}Dev{~~} & {~~}Test{~~} & {~~}Dev{~~} & {~~}Test{~~} & per epoch & per 10k samples \\
\hline
RNN & & 43.2 & & 82.4 & - & - \\
RNTN & & 45.7 & & 85.4 & 1657 & 1939 \\
DRNN & & 49.8 & & 86.8 & 431 & 504 \\
RLSTM & & \add{51.0} & & \add{88.0} & 140 & 164 \\
\hline
DCNN & & 48.5 & & 86.9 & - & - \\
CNN-MC & & 47.4 & & 88.1 & 2452 & 156 \\
\add{CNN} & \add{48.8} & \add{47.2} & \add{85.7} & \add{86.2} & 32 & 37 \\
\hline
PVEC & & 48.7 & & 87.8 & - & -\\
DAN & & 48.2 & & 86.8 & 73 & 5 \\
SVM & 40.1 & 38.3 & 78.6 & 81.3 & - & - \\
NBoW & 45.1 & 44.5 & 80.7 & 82.0 & 1 & 1 \\
\hline
\textbf{Ours} & 49.5 & 50.6 & 87.0 & 87.0 & 28 & 33 \\
$\quad$+ phrase labels & 53.4 & \textbf{51.2} & 88.9 & \textbf{88.6} & 445 & 28 \\
\hline
\end{tabular}
\caption{Comparison between our model and other baseline methods on Stanford
Sentiment Treebank. The top block lists recursive neural network models, the
second block are convolutional network models and the third block contains other
baseline methods, including the paragraph-vector model~\cite{le2014distributed},
the deep averaging network model~\cite{iyyer2015} and our implementation of
neural bag-of-words. The training time of baseline methods is taken
from~\cite{iyyer2015} or directly from the authors.  For our implementations,
timings were performed on a single core of a 2.6GHz Intel i7
processor.}
\label{table:sentiment}
\end{table*}

\paragraph{Datasets} We evaluate our model on sentence sentiment classification
task and news categorization task. For sentiment classification, we use the
Stanford Sentiment Treebank benchmark~\cite{socher2013sentiment}. The dataset
consists of 11855 parsed English sentences annotated \add{at both the root (i.e.
sentence) level and the phrase level using 5-class fine-grained labels.} We use
the standard 8544/1101/2210 split for training, development and testing
respectively. Following previous work, we also evaluate our model on the binary
classification variant of this benchmark, ignoring all neutral sentences. The
binary version has 6920/872/1821 sentences for training, development and
testing.

For the news categorization task, we evaluate on Sogou Chinese news
corpora.\footnote{\url{http://www.sogou.com/labs/dl/c.html}} The dataset
contains 10 different news categories in total, including Finance, Sports,
Technology and Automobile etc.
We use 79520 documents for training, 9940 for development and 9940 for testing.
To obtain Chinese word boundaries, we use
LTP-Cloud\footnote{\url{http://www.ltp-cloud.com/intro/en/}\\ $~\quad\ \ \
$\url{https://github.com/HIT-SCIR/ltp}}, an open-source Chinese NLP platform.

\paragraph{Baselines} \add{We implement the standard \textbf{SVM} method and the
neural bag-of-words model \textbf{NBoW} as baseline methods in both tasks. To
assess the proposed tensor-based feature map, we also implement a convolutional
neural network model \textbf{CNN} by replacing our filter with traditional
linear filter. The rest of the framework (such as feature averaging and
concatenation) remains the same.}

In addition, we compare our model with a wide range of top-performing models on
the sentence sentiment classification task. Most of these models fall into
either the category of recursive neural networks (RNNs) or the category of
convolutional neural networks (CNNs). The recursive neural network baselines
include standard \textbf{RNN}~\cite{socher2011semi}, \textbf{RNTN} with a small
core tensor in the composition function~\cite{socher2013sentiment}, the deep
recursive model \textbf{DRNN}~\cite{irsoy2014} and the most recent recursive
model using long-short-term-memory units \textbf{RLSTM}~\cite{tai2015improved}.
\add{These recursive models assume the input sentences are represented as parse
trees. As a benefit, they can readily utilize annotations at the phrase level.} In contrast, convolutional neural networks are
trained on sequence-level, taking the original sequence and its label as
training input. Such convolutional baselines include the dynamic CNN with k-max
pooling \textbf{DCNN}~\cite{kalchbrenner2014} and the convolutional model with
multi-channel \textbf{CNN-MC} by~\newcite{Kim14}. 
\add{To leverage the phrase-level annotations in the Stanford Sentiment
Treebank, all phrases and the corresponding labels are added as separate
instances when training the sequence models. We follow this strategy and report
results with and without phrase annotations.}



\paragraph{Word vectors} The word vectors are pre-trained on much larger
unannotated corpora to achieve better generalization given limited amount of
training data~\cite{Turian:2010}.  In particular, \add{for the English sentiment
classification task}, we use the publicly available 300-dimensional GloVe word
vectors trained on the Common Crawl with 840B tokens~\cite{pennington2014glove}.
\add{This choice of word vectors follows most recent work, such as \textbf{DAN}~\cite{iyyer2015} and \textbf{RLSTM}~\cite{tai2015improved}.} \add{For
Chinese news categorization, there is no widely-used publicly available word
vectors. Therefore,} we run \texttt{word2vec}~\cite{mikolov2013} to train
200-dimensional word vectors on the 1.6 million Chinese news articles. Both word
vectors are normalized to unit norm (i.e. $\|w\|_2^2=1$) and are fixed in the
experiments without fine-tuning.

\paragraph{Hyperparameter setting} We perform an extensive search on the
hyperparameters of our full model, our implementation of the \textbf{CNN} model
(with linear filters), and the \textbf{SVM} baseline.  For our model and the
\textbf{CNN} model, the initial learning rate of AdaGrad is fixed to 0.01 for
sentiment classification and 0.1 for news categorization, and the L2
regularization weight is fixed to $1e-5$ and $1e-6$ respectively based on
preliminary runs. The rest of the hyperparameters are randomly chosen as
follows: number of feature-mapping layers $\in \{1,2,3\}$, n-gram order $n\in
\{2,3\}$, hidden feature dimension $h\in \{50, 100, 200\}$, dropout probability
$\in \{0.0, 0.1, 0.3, 0.5\}$, and length decay $\lambda \in \{0.0, 0.3, 0.5\}$.
We run each configuration 3 times to explore different random initializations.
For the \textbf{SVM} baseline, we tune L2 regularization weight $C \in \{0.01,
0.1, 1.0, 10.0\}$, word cut-off frequency $\in \{1,2,3,5\}$ (i.e. pruning words
appearing less than this times) and n-gram feature order $n\in \{1,2,3\}$.

\paragraph{Implementation details} The source code is implemented in Python using the
Theano library~\cite{bergstra+al:2010-scipy}, a flexible linear algebra compiler that can
optimize user-specified computations (models) with efficient automatic low-level
implementations, including (back-propagated) gradient calculation. 

\section{Results}

\subsection{Overall Performance}

Table~\ref{table:sentiment} presents the performance of our model and other
baseline methods on Stanford Sentiment Treebank benchmark. \add{ Our full model
obtains the highest accuracy on both the development and test sets.
Specifically, it achieves 51.2\% and 88.6\% test accuracies on fine-grained and
binary tasks respectively\footnote{Best hyperparameter configuration based on
dev accuracy: 3 layers, 3-gram tensors (n=3), feature dimension $d=200$ and
length decay $\lambda=0.5$}. As shown in Table~\ref{table:avgstd}, our model
performance is relatively stable -- it remains high accuracies with around 0.5\%
standard deviation under different initializations and dropout rates.}

\add{Our full model is also several times faster than other top-performing
models.  For example, the convolutional model with multi-channel
(\textbf{CNN-MC}) runs over 2400 seconds per training epoch. In contrast, our
full model (with 3 feature layers) runs on average 28 seconds with only root
labels and on average 445 seconds with all labels.}

\add{Our results also show that the \textbf{CNN} model, where our feature map is
replaced with traditional linear map, performs worse than our full model.  This
observation confirms the importance of the proposed non-linear, tensor-based
feature mapping. The \textbf{CNN} model also lags behind the \textbf{DCNN} and
\textbf{CNN-MC} baselines, since the latter two propose several advancements
over standard CNN.}


\begin{table}[t]
\centering
\begin{tabular}{l|c|c}
\cline{2-3}
& \textbf{Dataset} & \textbf{Accuracy}\\
\hline
\multirow{2}{*}{Fine-grained} & Dev & 52.5 ($\pm$0.5) \% \\
& Test & 51.4 ($\pm$0.6) \% \\
\hline
\multirow{2}{*}{Binary} & Dev & 88.4 ($\pm$0.3) \% \\
& Test & 88.4 ($\pm$0.5) \%\\
\hline
\end{tabular}
\caption{Analysis of average accuracy and standard deviation of our model on
sentiment classification task.}
\label{table:avgstd}
\end{table}

Table~\ref{table:news} reports the results of \textbf{SVM}, \textbf{NBoW} and
our model on the news categorization task. Since the dataset is much larger compared
to the sentiment dataset (80K documents vs.  8.5K sentences), \add{the
\textbf{SVM} method is a competitive baseline. It achieves 78.5\% accuracy
compared to 74.4\% and 79.2\% obtained by the neural bag-of-words model and CNN
model}. In contrast, our model obtains 80.0\% accuracy on both the development and
test sets, outperforming the three baselines by a \add{0.8\% absolute margin}.
The best hyperparameter configuration in this task uses less feature layers and
lower n-gram order (specifically, 2 layers and $n=2$) compared to the sentiment
classification task. We hypothesize that the difference is due to the nature of the
two tasks: the document classification task requires to handle less compositions
or context interactions than sentiment analysis.

\begin{table}
\centering
\begin{tabular}{l|c|c}
\hline
\textbf{Model} & \textbf{Dev Acc.} & \textbf{Test Acc.} \\
\hline
SVM (1-gram) & 77.5 & 77.4 \\
SVM (2-gram) & 78.2 & 78.0 \\
SVM (3-gram) & 78.2 & 78.5 \\
NBoW & 74.4 & 74.4 \\
CNN & 79.5 & 79.2 \\
\textbf{Ours} & 80.0 & 80.0 \\
\hline
\end{tabular}
\caption{Performance of various methods on Chinese news categorization task. Our
model obtains better results than the SVM, NBoW and traditional CNN baselines.}
\label{table:news}
\end{table}

\subsection{Hyperparameter Analysis} We next investigate the impact of
hyperparameters in our model performance. We use the models trained on
fine-grained sentiment classification task with only root labels.

\begin{figure}[t]
\centering
\includegraphics[width=2.8in]{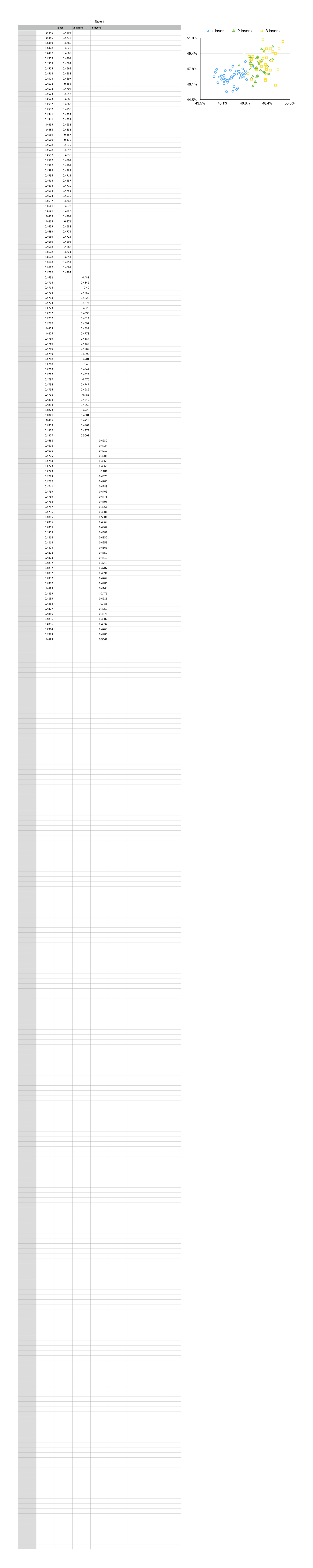}
\caption{Dev accuracy (x-axis) and test accuracy (y-axis) of independent runs of
our model on fine-grained sentiment classification task. Deeper architectures
achieve better accuracies.}
\label{figure:layers}
\end{figure}

\begin{figure}[t!]
\centering
\includegraphics[width=2.8in]{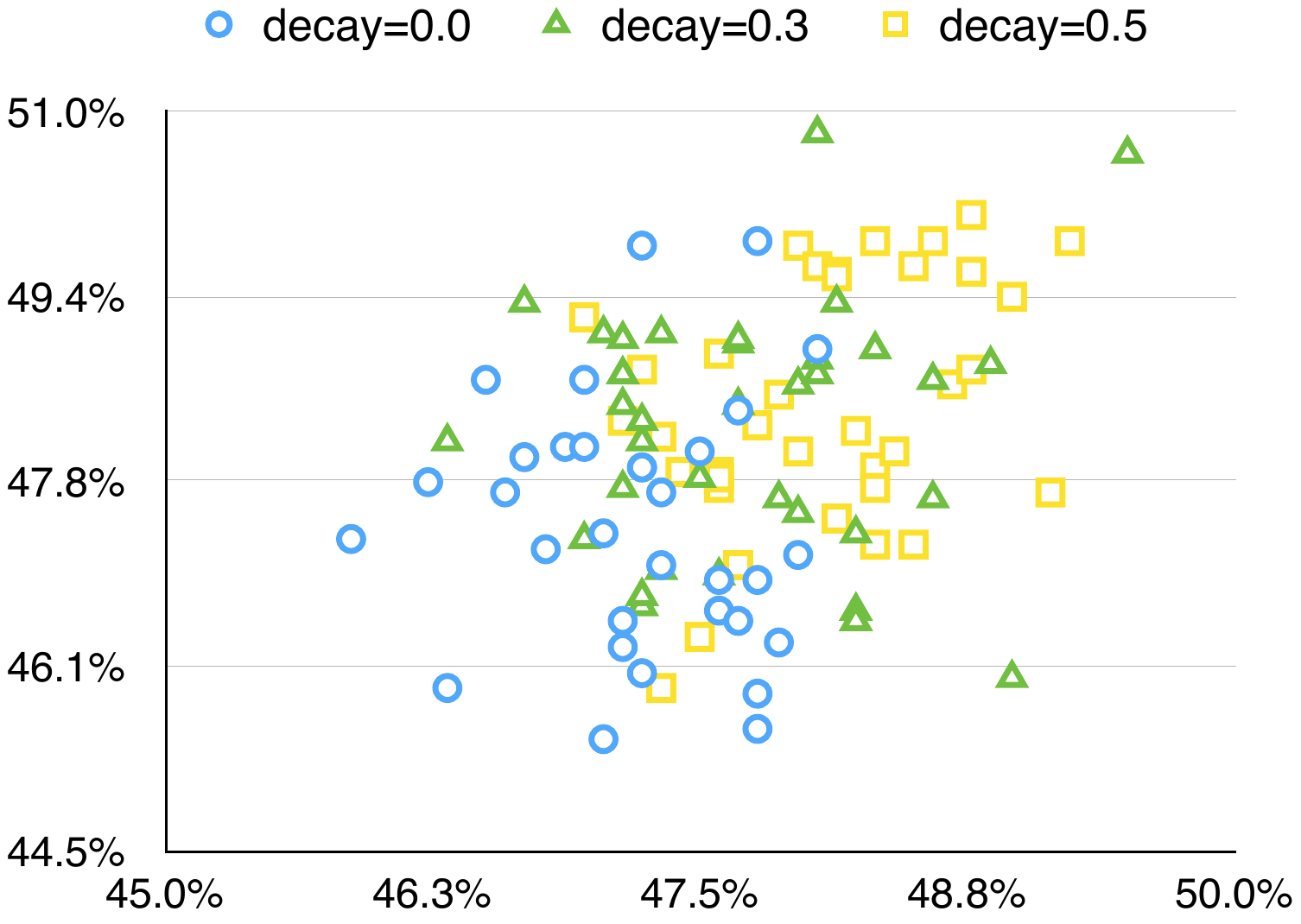}
\caption{Comparison of our model variations in sentiment classification task
when considering consecutive n-grams only (decaying factor $\lambda=0$) and when
considering non-consecutive n-grams ($\lambda > 0$). Modeling non-consecutive
n-gram features leads to better performance.}
\label{figure:decays}
\end{figure}

\begin{figure}[t]
\centering
\includegraphics[width=2.8in]{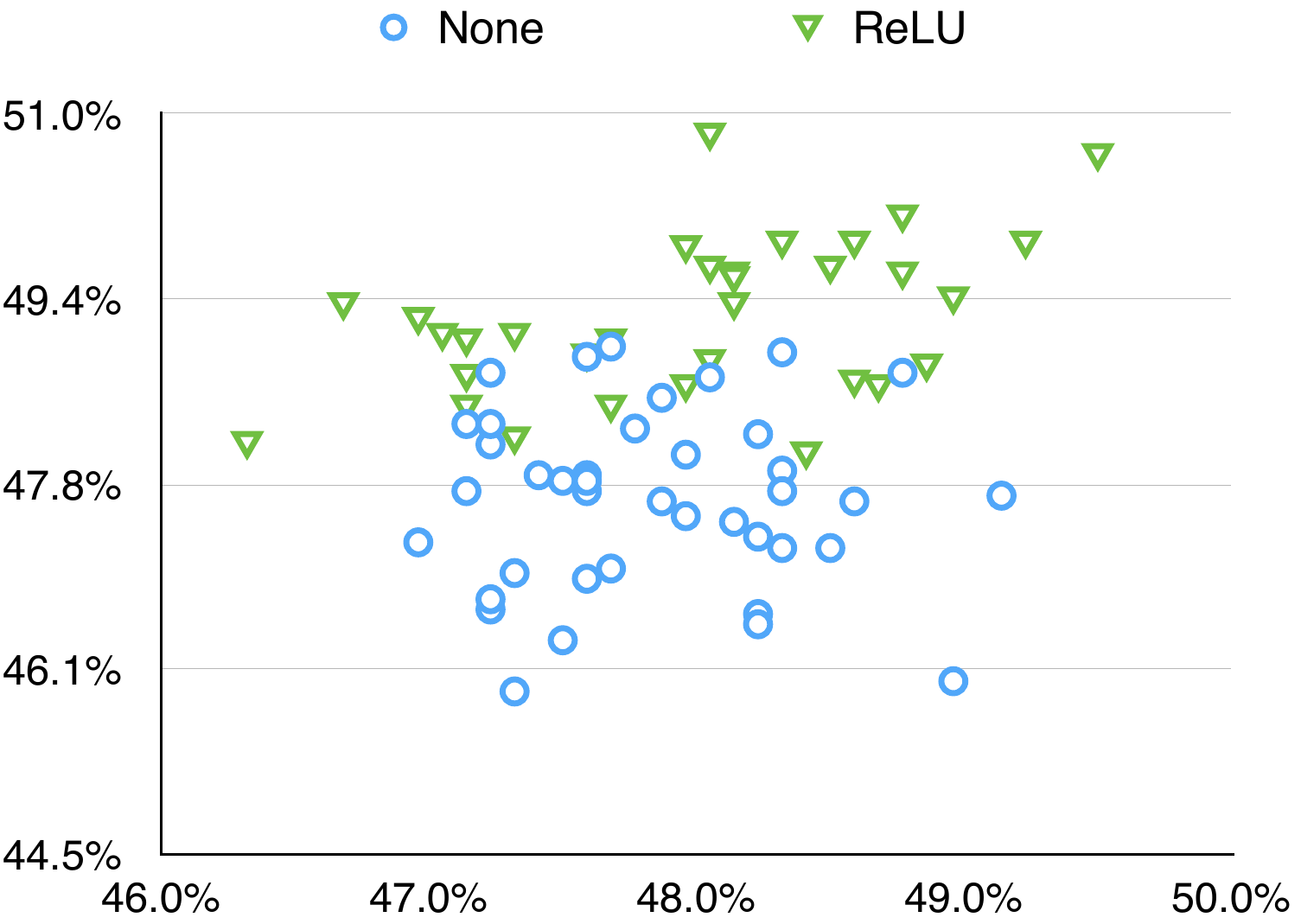}
\caption{Applying ReLU activation on top of tensor-based feature mapping leads
to better performance in sentiment classification task. Runs with no activation
are plotted as blue circles.}
\label{figure:activations}
\end{figure}

\begin{figure*}[t]
\centering
\begin{tabular}{ccccc}
\multicolumn{1}{>{\hspace{-4pt}}c<{\hspace{-4pt}}|}{\includegraphics[height=0.78in]{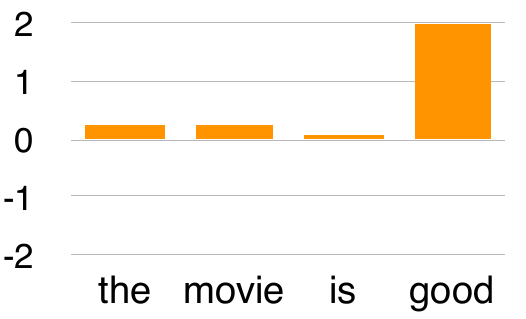}}
& \multicolumn{2}{>{\hspace{-4pt}}c<{\hspace{-4pt}}|}{\includegraphics[height=0.78in]{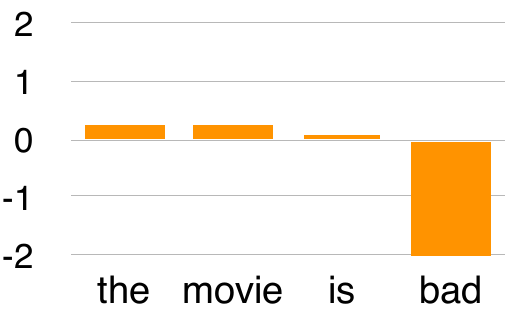}}
& \multicolumn{1}{>{\hspace{-4pt}}c<{\hspace{-4pt}}|}{\includegraphics[height=0.78in]{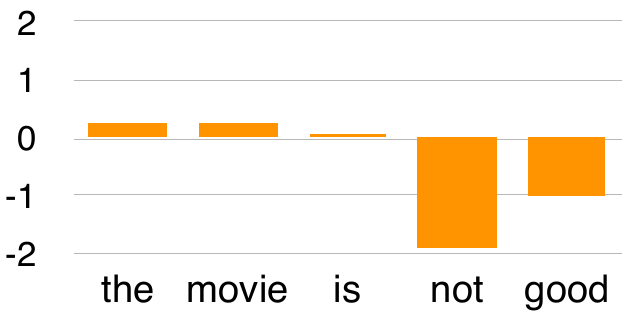}}
& \includegraphics[height=0.78in]{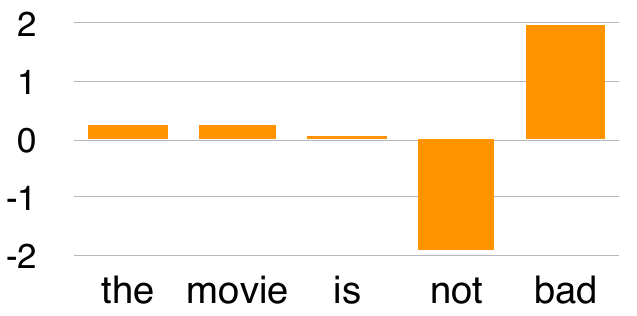} \\
(1) positive prediction & \multicolumn{2}{c}{(2) negative prediction} & (3) negative prediction & (4)
positive prediction \\
\\
\multicolumn{2}{c|}{\includegraphics[height=0.78in]{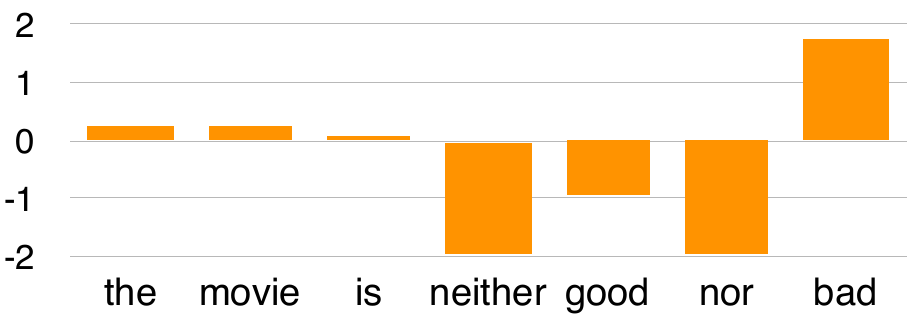}} 
& \multicolumn{3}{>{\hspace{-8pt}}c}{\includegraphics[height=0.78in]{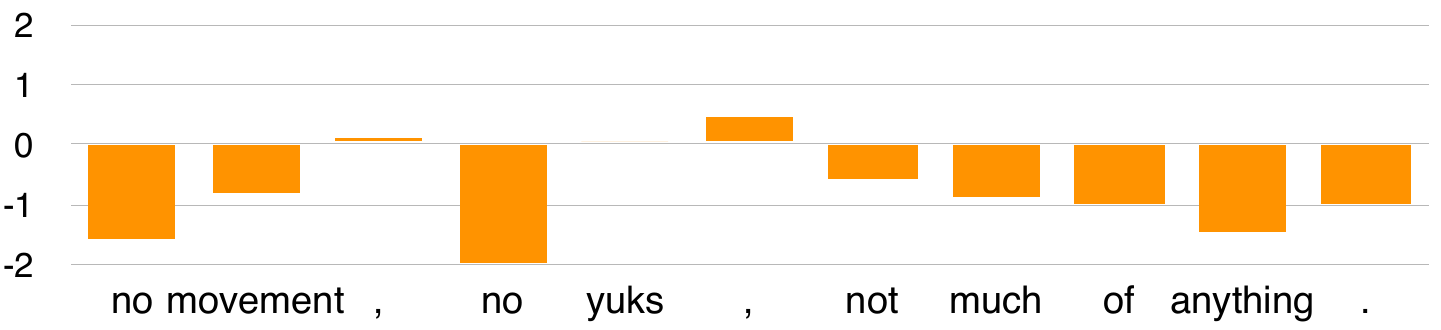}} \\
\multicolumn{2}{c}{(5) negative prediction}
& \multicolumn{3}{c}{(6) negative prediction (ground truth: negative)}  \\
\\
\multicolumn{5}{c}{\includegraphics[height=0.78in]{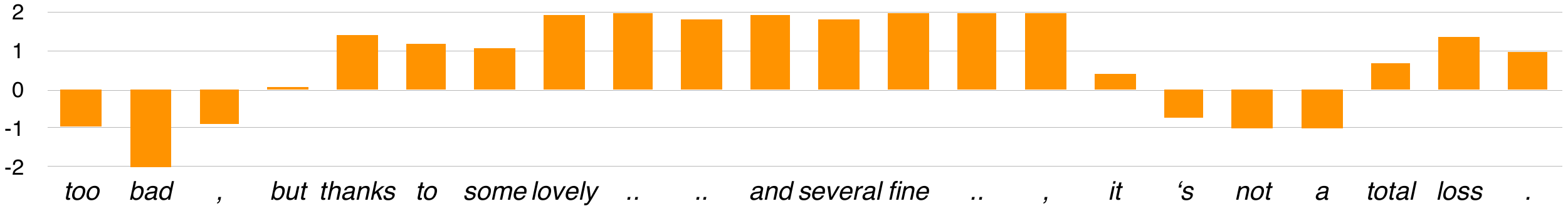}} \\
\multicolumn{5}{c}{(7) positive prediction (ground truth: positive)} \\
\end{tabular}
\caption{Example sentences and their sentiments predicted by our model trained
with root labels.
The predicted sentiment scores at each word position are plotted. Examples
(1)-(5) are synthetic inputs, (6) and (7) are two real inputs from the test set.
Our model successfully identifies negation, double negation and phrases with
different sentiment in one sentence.}
\label{figure:examples}
\end{figure*}

\paragraph{Number of layers} We plot the fine-grained sentiment classification
accuracies obtained during hyperparameter grid search.
Figure~\ref{figure:layers} illustrates how the number of feature layers impacts
the model performance. As shown in the figure, adding higher-level features
clearly improves the classification accuracy across various hyperparameter
settings and initializations.

\paragraph{Non-consecutive n-gram features} We also analyze the effect of
modeling non-consecutive n-grams.
Figure~\ref{figure:decays} splits the model accuracies according to the choice
of span decaying factor $\lambda$. Note when $\lambda=0$, the model applies
feature extractions to consecutive n-grams only. As shown in
Figure~\ref{figure:decays}, this setting leads to consistent performance drop. This result confirms the importance of handling
non-consecutive n-gram patterns.

\paragraph{Non-linear activation} Finally, we verify the
effectiveness of rectified linear unit activation function (ReLU) by comparing
it with no activation (or identity activation $f(x)=x$). As
shown in Figure~\ref{figure:activations}, our model with ReLU activation
generally outperforms its variant without ReLU. The observation is consistent
with previous work on convolutional neural networks and other neural network
models.

\subsection{Example Predictions}
Figure~\ref{figure:examples} gives examples of input sentences and the
corresponding predictions of our model in fine-grained sentiment classification.
To see how our model captures the sentiment at different local context, we apply
the learned softmax activation to the extracted features at each position
without taking the average. That is, for each index $i$, we obtain the local
sentiment $p=\text{softmax}\left(\mathbf{W}^\top \left(z^{(1)}[i]\oplus
z^{(2)}[i]\oplus z^{(3)}[i]\right)\right)$. We plot the expected sentiment
scores $\sum_{s=-2}^2 s\cdot p(s)$, where a score of 2 means ``very positive'',
0 means ``neutral'' and -2 means ``very negative''. As shown in the figure, our
model successfully learns negation and double negation. The model also
identifies positive and negative segments appearing in the sentence.

\section{Conclusion}
We proposed a feature mapping operator for convolutional neural networks by modeling n-gram interactions based on tensor product and evaluating all non-consecutive n-gram vectors. The associated parameters are maintained as a low-rank tensor, which leads to efficient feature extraction via dynamic programming. The model achieves top performance on standard sentiment classification and document categorization tasks.
\section*{Acknowledgments}
\add{We thank Kai Sheng Tai, Mohit Iyyer and Jordan Boyd-Graber for answering
questions about their paper. We also thank Yoon Kim, the MIT NLP group and the
reviewers for their comments. We acknowledge the support of the U.S.  Army
Research Office under grant number W911NF-10-1-0533. The work is developed in
collaboration with the Arabic Language Technologies (ALT) group at Qatar
Computing Research Institute (QCRI).  Any opinions, findings, conclusions, or
recommendations expressed in this paper are those of the authors, and do not
necessarily reflect the views of the funding organizations.}

\bibliography{paper}

\begin{thebibliography}{}

\bibitem[\protect\citename{Bahdanau \bgroup et al.\egroup
  }2014]{bahdanau2014neural}
Dzmitry Bahdanau, Kyunghyun Cho, and Yoshua Bengio.
\newblock 2014.
\newblock Neural machine translation by jointly learning to align and
  translate.
\newblock {\em arXiv preprint arXiv:1409.0473}.

\bibitem[\protect\citename{Bengio \bgroup et al.\egroup
  }2003]{bengio2003neural}
Yoshua Bengio, R{\'e}jean Ducharme, Pascal Vincent, and Christian Janvin.
\newblock 2003.
\newblock A neural probabilistic language model.
\newblock {\em The Journal of Machine Learning Research}, 3:1137--1155.

\bibitem[\protect\citename{Bergstra \bgroup et al.\egroup
  }2010]{bergstra+al:2010-scipy}
James Bergstra, Olivier Breuleux, Fr{\'{e}}d{\'{e}}ric Bastien, Pascal Lamblin,
  Razvan Pascanu, Guillaume Desjardins, Joseph Turian, David Warde-Farley, and
  Yoshua Bengio.
\newblock 2010.
\newblock Theano: a {CPU} and {GPU} math expression compiler.
\newblock In {\em Proceedings of the Python for Scientific Computing Conference
  ({SciPy})}.

\bibitem[\protect\citename{Chen and Manning}2014]{chen2014fast}
Danqi Chen and Christopher~D Manning.
\newblock 2014.
\newblock A fast and accurate dependency parser using neural networks.
\newblock In {\em Proceedings of the 2014 Conference on Empirical Methods in
  Natural Language Processing (EMNLP)}, pages 740--750.

\bibitem[\protect\citename{Collobert and Weston}2008]{Collobert:2008}
R.~Collobert and J.~Weston.
\newblock 2008.
\newblock A unified architecture for natural language processing: Deep neural
  networks with multitask learning.
\newblock In {\em International Conference on Machine Learning, {ICML}}.

\bibitem[\protect\citename{Collobert \bgroup et al.\egroup
  }2011]{collobert2011natural}
Ronan Collobert, Jason Weston, L{\'e}on Bottou, Michael Karlen, Koray
  Kavukcuoglu, and Pavel Kuksa.
\newblock 2011.
\newblock Natural language processing (almost) from scratch.
\newblock {\em The Journal of Machine Learning Research}, 12:2493--2537.

\bibitem[\protect\citename{Devlin \bgroup et al.\egroup }2014]{devlin2014fast}
Jacob Devlin, Rabih Zbib, Zhongqiang Huang, Thomas Lamar, Richard Schwartz, and
  John Makhoul.
\newblock 2014.
\newblock Fast and robust neural network joint models for statistical machine
  translation.
\newblock In {\em 52nd Annual Meeting of the Association for Computational
  Linguistics}.

\bibitem[\protect\citename{Duchi \bgroup et al.\egroup }2011]{adagrad}
John Duchi, Elad Hazan, and Yoram Singer.
\newblock 2011.
\newblock Adaptive subgradient methods for online learning and stochastic
  optimization.
\newblock {\em The Journal of Machine Learning Research}, 12:2121--2159.

\bibitem[\protect\citename{Gao \bgroup et al.\egroup }2014]{gao2014modeling}
Jianfeng Gao, Patrick Pantel, Michael Gamon, Xiaodong He, Li~Deng, and Yelong
  Shen.
\newblock 2014.
\newblock Modeling interestingness with deep neural networks.
\newblock In {\em Proceedings of the 2013 Conference on Empirical Methods in
  Natural Language Processing}.

\bibitem[\protect\citename{Hinton \bgroup et al.\egroup }2012]{dropout}
Geoffrey~E Hinton, Nitish Srivastava, Alex Krizhevsky, Ilya Sutskever, and
  Ruslan~R Salakhutdinov.
\newblock 2012.
\newblock Improving neural networks by preventing co-adaptation of feature
  detectors.
\newblock {\em arXiv preprint arXiv:1207.0580}.

\bibitem[\protect\citename{Irsoy and Cardie}2014]{irsoy2014}
Ozan Irsoy and Claire Cardie.
\newblock 2014.
\newblock Deep recursive neural networks for compositionality in language.
\newblock In {\em Advances in Neural Information Processing Systems}.

\bibitem[\protect\citename{Iyyer \bgroup et al.\egroup }2015]{iyyer2015}
Mohit Iyyer, Varun Manjunatha, Jordan Boyd-Graber, and Hal~Daume III.
\newblock 2015.
\newblock Deep unordered composition rivals syntactic methods for text
  classification.
\newblock In {\em Association for Computational Linguistics}.

\bibitem[\protect\citename{Kalchbrenner and Blunsom}2013]{kalchbrennerB13}
Nal Kalchbrenner and Phil Blunsom.
\newblock 2013.
\newblock Recurrent continuous translation models.
\newblock In {\em Proceedings of the 2013 Conference on Empirical Methods in
  Natural Language Processing ({EMNLP} 2013)}, pages 1700--1709.

\bibitem[\protect\citename{Kalchbrenner \bgroup et al.\egroup
  }2014]{kalchbrenner2014}
Nal Kalchbrenner, Edward Grefenstette, and Phil Blunsom.
\newblock 2014.
\newblock A convolutional neural network for modelling sentences.
\newblock In {\em Proceedings of the 52th Annual Meeting of the Association for
  Computational Linguistics}.

\bibitem[\protect\citename{Kartsaklis \bgroup et al.\egroup
  }2012]{kartsaklis2012unified}
Dimitri Kartsaklis, Mehrnoosh Sadrzadeh, and Stephen Pulman.
\newblock 2012.
\newblock A unified sentence space for categorical distributional-compositional
  semantics: Theory and experiments.
\newblock In {\em In Proceedings of COLING: Posters}.

\bibitem[\protect\citename{Kim}2014]{Kim14}
Yoon Kim.
\newblock 2014.
\newblock Convolutional neural networks for sentence classification.
\newblock In {\em Proceedings of the Empiricial Methods in Natural Language
  Processing (EMNLP 2014)}.

\bibitem[\protect\citename{K{\"u}chler and Goller}1996]{kuchler1996inductive}
Andreas K{\"u}chler and Christoph Goller.
\newblock 1996.
\newblock Inductive learning in symbolic domains using structure-driven
  recurrent neural networks.
\newblock In {\em KI-96: Advances in Artificial Intelligence}, pages 183--197.

\bibitem[\protect\citename{Le and Mikolov}2014]{le2014distributed}
Quoc Le and Tomas Mikolov.
\newblock 2014.
\newblock Distributed representations of sentences and documents.
\newblock In {\em Proceedings of the 31st International Conference on Machine
  Learning (ICML-14)}, pages 1188--1196.

\bibitem[\protect\citename{Le and Zuidema}2015]{le2015compositional}
Phong Le and Willem Zuidema.
\newblock 2015.
\newblock Compositional distributional semantics with long short term memory.
\newblock In {\em Proceedings of Joint Conference on Lexical and Computational
  Semantics (*SEM)}.

\bibitem[\protect\citename{LeCun \bgroup et al.\egroup }1998]{lecun-98}
Y.~LeCun, L.~Bottou, Y.~Bengio, and P.~Haffner.
\newblock 1998.
\newblock Gradient-based learning applied to document recognition.
\newblock {\em Proceedings of the IEEE}, 86(11):2278--2324, November.

\bibitem[\protect\citename{Lei \bgroup et al.\egroup }2014]{Lei14}
Tao Lei, Yu~Xin, Yuan Zhang, Regina Barzilay, and Tommi Jaakkola.
\newblock 2014.
\newblock Low-rank tensors for scoring dependency structures.
\newblock In {\em Proceedings of the 52th Annual Meeting of the Association for
  Computational Linguistics}. Association for Computational Linguistics.

\bibitem[\protect\citename{Mikolov \bgroup et al.\egroup
  }2010]{mikolov2010recurrent}
Tomas Mikolov, Martin Karafi{\'a}t, Lukas Burget, Jan Cernock{\`y}, and Sanjeev
  Khudanpur.
\newblock 2010.
\newblock Recurrent neural network based language model.
\newblock In {\em INTERSPEECH 2010, 11th Annual Conference of the International
  Speech Communication Association, Makuhari, Chiba, Japan, September 26-30,
  2010}, pages 1045--1048.

\bibitem[\protect\citename{Mikolov \bgroup et al.\egroup }2013]{mikolov2013}
Tomas Mikolov, Kai Chen, Greg Corrado, and Jeffrey Dean.
\newblock 2013.
\newblock Efficient estimation of word representations in vector space.
\newblock {\em CoRR}.

\bibitem[\protect\citename{Mitchell and Lapata}2008]{mitchell2008vector}
Jeff Mitchell and Mirella Lapata.
\newblock 2008.
\newblock Vector-based models of semantic composition.
\newblock In {\em ACL}, pages 236--244.

\bibitem[\protect\citename{Pennington \bgroup et al.\egroup
  }2014]{pennington2014glove}
Jeffrey Pennington, Richard Socher, and Christopher~D Manning.
\newblock 2014.
\newblock Glove: Global vectors for word representation.
\newblock volume~12.

\bibitem[\protect\citename{Pollack}1990]{pollack1990recursive}
Jordan~B Pollack.
\newblock 1990.
\newblock Recursive distributed representations.
\newblock {\em Artificial Intelligence}, 46:77--105.

\bibitem[\protect\citename{Shen \bgroup et al.\egroup }2014]{shen2014learning}
Yelong Shen, Xiaodong He, Jianfeng Gao, Li~Deng, and Gr{\'e}goire Mesnil.
\newblock 2014.
\newblock Learning semantic representations using convolutional neural networks
  for web search.
\newblock In {\em Proceedings of the companion publication of the 23rd
  international conference on World wide web companion}, pages 373--374.
  International World Wide Web Conferences Steering Committee.

\bibitem[\protect\citename{Socher \bgroup et al.\egroup
  }2011a]{SocherEtAl2011:RNN}
Richard Socher, Cliff~C. Lin, Andrew~Y. Ng, and Christopher~D. Manning.
\newblock 2011a.
\newblock Parsing natural scenes and natural language with recursive neural
  networks.
\newblock In {\em Proceedings of the 26th International Conference on Machine
  Learning (ICML)}.

\bibitem[\protect\citename{Socher \bgroup et al.\egroup }2011b]{socher2011semi}
Richard Socher, Jeffrey Pennington, Eric~H Huang, Andrew~Y Ng, and
  Christopher~D Manning.
\newblock 2011b.
\newblock Semi-supervised recursive autoencoders for predicting sentiment
  distributions.
\newblock In {\em Proceedings of the Conference on Empirical Methods in Natural
  Language Processing}, pages 151--161. Association for Computational
  Linguistics.

\bibitem[\protect\citename{Socher \bgroup et al.\egroup
  }2013]{socher2013sentiment}
Richard Socher, Alex Perelygin, Jean Wu, Jason Chuang, Christopher~D. Manning,
  Andrew~Y. Ng, and Christopher Potts.
\newblock 2013.
\newblock Recursive deep models for semantic compositionality over a sentiment
  treebank.
\newblock In {\em Proceedings of the 2013 Conference on {E}mpirical {M}ethods
  in {N}atural {L}anguage {P}rocessing}, pages 1631--1642, October.

\bibitem[\protect\citename{Sutskever \bgroup et al.\egroup
  }2014]{sutskever2014sequence}
Ilya Sutskever, Oriol Vinyals, and Quoc~VV Le.
\newblock 2014.
\newblock Sequence to sequence learning with neural networks.
\newblock In {\em Advances in Neural Information Processing Systems}, pages
  3104--3112.

\bibitem[\protect\citename{Tai \bgroup et al.\egroup }2015]{tai2015improved}
Kai~Sheng Tai, Richard Socher, and Christopher~D Manning.
\newblock 2015.
\newblock Improved semantic representations from tree-structured long
  short-term memory networks.
\newblock In {\em Proceedings of the 53th Annual Meeting of the Association for
  Computational Linguistics}.

\bibitem[\protect\citename{Turian \bgroup et al.\egroup }2010]{Turian:2010}
Joseph Turian, Lev Ratinov, and Yoshua Bengio.
\newblock 2010.
\newblock Word representations: A simple and general method for semi-supervised
  learning.
\newblock In {\em Proceedings of the 48th Annual Meeting of the Association for
  Computational Linguistics}, ACL '10. Association for Computational
  Linguistics.

\bibitem[\protect\citename{Yih \bgroup et al.\egroup }2014]{yih2014semantic}
Wen-tau Yih, Xiaodong He, and Christopher Meek.
\newblock 2014.
\newblock Semantic parsing for single-relation question answering.
\newblock In {\em Proceedings of ACL}.

\bibitem[\protect\citename{Zhang and LeCun}2015]{zhang2015text}
Xiang Zhang and Yann LeCun.
\newblock 2015.
\newblock Text understanding from scratch.
\newblock {\em arXiv preprint arXiv:1502.01710}.

\end{thebibliography}
\bibliographystyle{acl}

\end{document}